\documentclass[letterpaper, 10 pt, conference]{ieeeconf}  

\IEEEoverridecommandlockouts                              

\makeatletter
\let\NAT@parse\undefined
\makeatother

\usepackage{xcolor}    
\usepackage{amsmath}
\usepackage{graphicx}
\usepackage{cite}
\usepackage{amsfonts}
\usepackage{multirow}
\usepackage{bm}
\usepackage{array}
\usepackage[colorlinks=false]{hyperref}

\newif\ifanonymized
\anonymizedfalse   

\ifanonymized
    \title{\LARGE \bf
Push Anything: Single- and Multi-Object Pushing \\ From First Sight with Contact-Implicit MPC}
    \author{
        Anonymous Authors
    }
\else
    \title{\LARGE \bf
    Push Anything: Single- and Multi-Object Pushing \\ From First Sight with Contact-Implicit MPC
    \thanks{*Contributed equally to this work}
    \thanks{$^{1}$All authors are with the General Robotics, Automation, Sensing, and Perception (GRASP) Laboratory, University of Pennsylvania, USA \{xuanhien, yufeiyg, haorany8, ericcui, sidmody, bjacosta, stfelix, bibit, posa\}@seas.upenn.edu}
    \author{Hien Bui$^{*}$, Yufeiyang Gao$^{*}$, Haoran Yang$^{*}$, Eric Cui, Siddhant Mody, \\ Brian Acosta, Thomas Stephen Felix, Bibit Bianchini, and Michael Posa$^{1}$}}
\fi

\begin{document}

\maketitle
\thispagestyle{empty}
\pagestyle{empty}

\begin{abstract}
Non-prehensile manipulation of diverse objects remains a core challenge in robotics, driven by unknown physical properties and the complexity of contact-rich interactions.
Recent advances in contact-implicit model predictive control (CI-MPC), with contact reasoning embedded directly in the trajectory optimization, have shown promise in tackling the task efficiently and robustly.
However, demonstrations have been limited to narrowly curated examples.
In this work, we showcase the broader capabilities of CI-MPC through precise planar pushing tasks over a wide range of object geometries, including multi-object domains.
These scenarios demand reasoning over numerous inter-object and object-environment contacts to strategically manipulate and de-clutter the environment, which was intractable for prior CI-MPC methods.
To achieve this, we introduce Consensus Complementarity Control Plus (C3+), an enhanced CI-MPC algorithm integrated into a complete pipeline spanning object scanning, mesh reconstruction, and hardware execution.
Compared to its predecessor C3, C3+ achieves substantially faster solve times, enabling real-time performance even in multi-object pushing tasks.
On hardware, our system achieves overall 98\% success rate across 33 objects, reaching pose goals within tight tolerances.
The average time-to-goal is approximately 0.5, 1.6, 3.2, and 5.3 minutes for 1-, 2-, 3-, and 4-object tasks, respectively.
Project page: \url{https://dairlab.github.io/push-anything}.
\end{abstract}

\section{Introduction}
A key challenge in robotic manipulation is planning dynamic, contact-rich motions with objects of arbitrary geometries, especially within cluttered, multi-contact environments.
Model-based approaches like contact-implicit model predictive control (CI-MPC) \cite{Aydinoglu2024, cleac2024, kurtz2025} are promising for these tasks as they include contact terms as part of real-time trajectory optimization.
However, CI-MPC relies on local approximations of nonlinear dynamics, which can restrict its effectiveness to regions where these approximations hold.
To address this limitation, Venkatesh, Bianchini et al. \cite{Venkatesh2025} augment CI-MPC, specifically using Consensus Complementarity Control (C3) \cite{Aydinoglu2024}, with global exploration through low-dimensional sampling of end effector positions.
Their approach separates task execution into a contact-free stage, where the robot follows easily computed collision-free paths, and a contact-rich stage, during which CI-MPC effectively guides the robot to make and break contacts as long as the local dynamics permit progress toward the goal.
Despite these advances, prior demonstrations have largely been restricted to single-object scenarios with precisely known geometries, mass, and inertia from CAD models, limiting their applicability to online, real-world settings.
Moreover, tasks involving complex multi-object interactions, such as resolving cluttered scenes, remain intractable for prior CI-MPC methods as problem complexity grows exponentially with the number of contacts.

We introduce \textit{Push Anything}, a manipulation pipeline for real-time planar pushing of a wide variety of objects, including multi-object scenes.
Push Anything integrates real-world object scanning and mesh reconstruction, robust object tracking, and a controller built on the framework of \cite{Venkatesh2025} with improvements to the local CI-MPC.
Central to these improvements is Consensus Complementarity Control Plus (C3+), an enhanced version of C3 that substantially accelerates solve times, enabling the system to reason over numerous inter-object and object-environment contacts across multi-step horizons.
Our pipeline shows high-precision manipulation of diverse objects on hardware, including multi-object decluttering tasks that were previously intractable.

\begin{figure}[t]
    \centering
    \includegraphics[width=\linewidth]{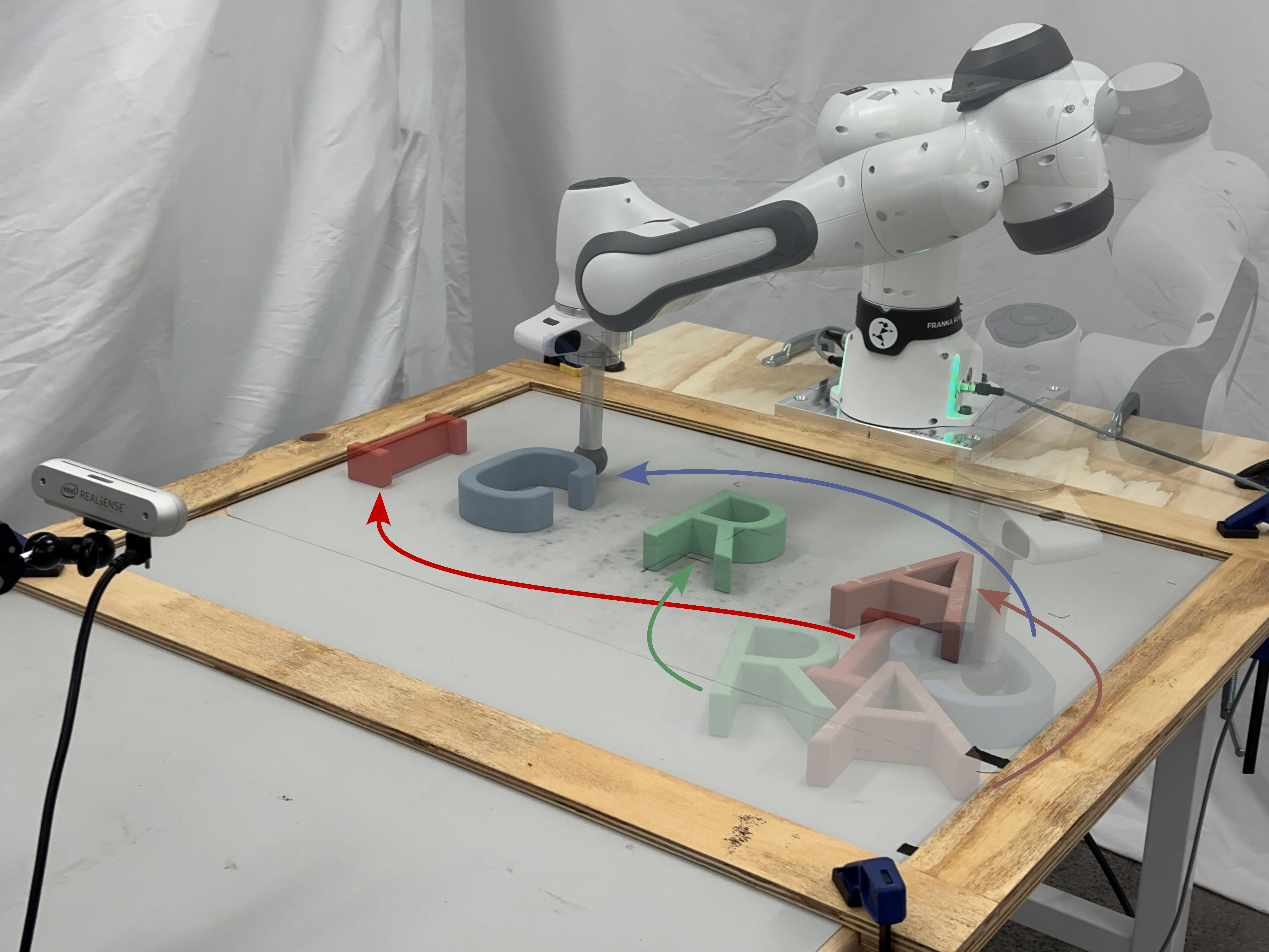}
    \caption{Experimental Setup: The Franka Emika Panda arm uses a spherical end effector to push and rearrange four objects from an initial cluttered configuration.}
	\label{img.exp_setup}
    \vspace{-0.5cm}
\end{figure}

\begin{figure*}[t]
    \centering
    \includegraphics[width=\linewidth]{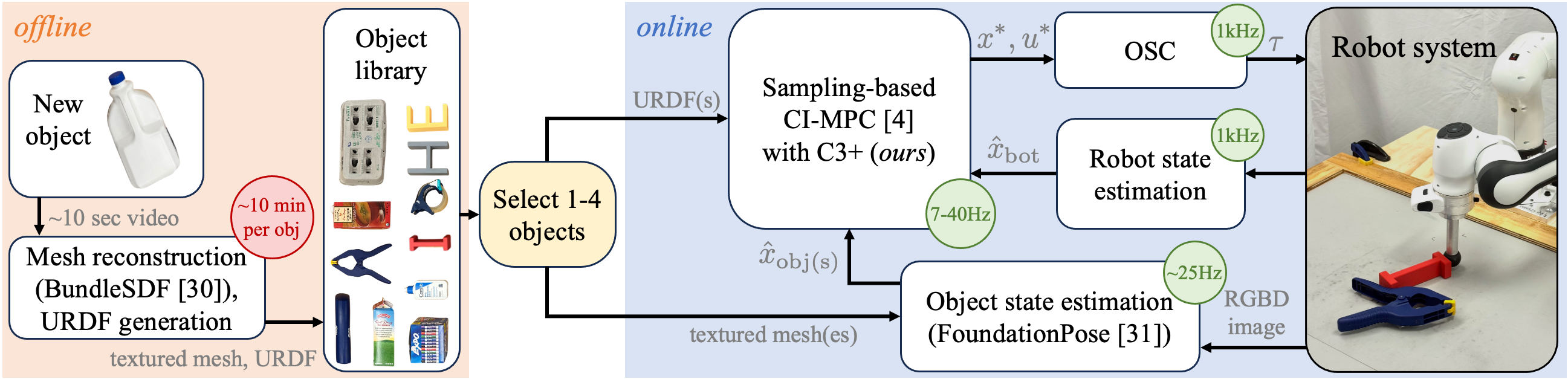}
    \caption{System diagram of the Push Anything framework.}
	\label{img.diagram}
    \vspace{-0.5cm}
\end{figure*}

In this paper, we contribute:
\begin{itemize}
    \item \textit{Push Anything}: A fully-integrated manipulation system that processes real-world scans to reconstruct object geometry, tracks objects robustly, and plans contact-rich pushing motions in real-time.
    \item Consensus Complementarity Control Plus (C3+): An enhanced CI-MPC algorithm capable of efficiently reasoning over numerous inter-object and object–environment contacts (demonstrated with up to 19 contact pairs), while planning over a multi-step horizon to enable precise multi-object manipulation.
    \item Hardware validation: Extensive real-world experiments demonstrating high-precision planar pushing over 928 trials with 98\% overall success rate.
\end{itemize}

\section{Related Work}
\subsection{Planar Pushing}
Non-prehensile planar pushing is a canonical problem in robotics, with a rich history rooted in model-based control.
Pioneering work established the mechanical foundations of pushing based on friction and contact geometry \cite{mason1986, mason1996, mason1998}, an analytical lineage that extended into planning combined pushing and grasping actions in cluttered environments \cite{ogarFrameworkPushGraspingClutter}, developing planar pushing interaction datasets \cite{kuan-ting2016}, and formulating novel contact models \cite{yiPreciseObjectSliding2023}.
While these methods offer precise control by explicitly modeling physics, they face significant scalability challenges.
The primary difficulty lies in the combinatorial complexity of hybrid dynamics; searching through all possible sequences of contact modes (e.g., sticking, slipping) is intractable for multi-object scenarios.

To sidestep modeling complexities, data-driven approaches have gained prominence.
Push-Net demonstrated robust repositioning of single objects using a recurrent policy learned from interaction history \cite{li2018}.
Building on this, reinforcement learning (RL) methods improved generalization and task complexity: multi-modal exploration captured hybrid dynamics to enable more accurate motions \cite{10341629}, while location-based attention allowed for goal-directed, collision-avoiding pushing in cluttered scenes \cite{dengler2025learninggoaldirectedobjectpushing}.
However, the learning paradigm introduces its own limitations. These methods are typically data-intensive and, to date, have been demonstrated primarily on single-object tasks, failing to solve the general problem of multi-object rearrangement.

\subsection{Contact-Implicit MPC}
CI-MPC enables online reasoning about contact-rich dynamics.
These approaches typically embed rigid-body contact physics directly into the predictive model, allowing control inputs and contact interactions to be optimized simultaneously without pre-specified mode schedules.
To achieve real-time performance, prior work relies on local approximations of the nonlinear dynamics.
For example, some approaches \cite{Aydinoglu2024, Yang2024, menager:hal-05201780} explicitly model non-smooth, multi-modal dynamics and employ ADMM-based consensus decomposition to parallelize contact scheduling.
Other methods \cite{cleac2024, jiang2024contactimplicitmodelpredictivecontrol, kurtz2025} use smoothed contact dynamics, enabling differentiable optimization and faster real-time performance, but at the cost of reduced contact fidelity and artifacts such as forces acting at a distance.
Because CI-MPC depends on these local approximations, its ability to reason about contacts in distant regions or escape local minima is limited.

\subsection{Sampling-Based MPC}
One prominent category of sampling-based MPC directly samples a multitude of control input sequences, a strategy popularized by methods like Model Predictive Path Integral control (MPPI) \cite{gradyMPPI} or Cross Entropy Method (CEM) \cite{reuvenCEM}.
These approaches use a forward dynamics model, often within a fast simulator, to roll out each sequence and evaluate its outcome against a cost function \cite{zhang2025, li2024drop, hess2024sampling, howell2022predictivesamplingrealtimebehaviour}.
The best-performing trajectory is then selected for execution. While effective for exploring the input space, this direct sampling approach can be computationally demanding and is often challenged by the curse of dimensionality, particularly for high-dimensional systems or long-horizon tasks.

To combine the broad exploration of sampling with the precision of trajectory optimization, another line of work integrates sampling with CI-MPC.
These hybrid methods typically use sampling to generate a diverse set of high-level candidate goals, thereby avoiding the local minima that can trap a standalone CI-MPC.
For instance, some methods sample low-dimensional end effector positions online \cite{Venkatesh2025}, while others leverage offline sampling of stable grasp configurations \cite{suh2025}.
For each candidate, local CI-MPC is solved to find a dynamically feasible trajectory.
The robot then executes the motion plan corresponding to the overall lowest-cost solution, effectively using CI-MPC as a powerful local planner within a broader sampling framework.
Our paper builds directly upon this line of work, particularly the framework by Venkatesh, Bianchini et al. \cite{Venkatesh2025}.

\section{Background}
We first introduce hybrid models for contact dynamics (\S \ref{subsec:bg_hybrid_models}) then describe the general formulation of CI-MPC (\S \ref{subsec:bg_cimpc}). Finally, we discuss how a sampling approach can extend CI-MPC toward more global solutions (\S \ref{subsec:bg_sampling_cimpc}).

\subsection{Hybrid Models for Contact Dynamics}\label{subsec:bg_hybrid_models}
Contact dynamics are inherently discontinuous, involving sticking, sliding, or separation.  
Hybrid models capture these behaviors by switching dynamics depending on the active contact mode.  
A compact representation for contact dynamics uses complementarity constraints:
\begin{subequations}
    \label{eqn:dynamics}
    \begin{align}
        x_{k+1} &= f(x_k, u_k, \lambda_k), \\
        0 \leq \lambda_k &\perp \Phi(x_k, u_k, \lambda_k) \geq 0,
    \end{align}
\end{subequations}
where $x_k \in \mathbb{R}^{n_x}$ is the state, $u_k \in \mathbb{R}^{n_u}$ the control input, $\lambda_k \in \mathbb{R}^{n_\lambda}$ contact forces, and $\Phi$ the signed distance (gap function) between potential contact pairs.
The active set $\Phi \geq 0$ defines the domain of each hybrid mode, while $f$ and $\Phi$ together implicitly specify the corresponding dynamics.
Simulating these dynamics requires solving (\ref{eqn:dynamics}b) for $\lambda_k$, a nonlinear complementarity problem (NCP) \cite{doi:10.1137/1.9781611970715}, and then updating $x_{k+1}$ via (\ref{eqn:dynamics}a).

\subsection{Contact-Implicit MPC}\label{subsec:bg_cimpc}
Given the contact dynamics in (\ref{eqn:dynamics}), the CI-MPC formulation \cite{doi:10.1177/0278364913506757, doi:10.1177/0278364919849235} treats $\lambda_k$ as decision variables within the optimization.
This increases the number of variables and constraints, but often leads to better-conditioned problems.
\begin{subequations}
\label{eqn:ci_mpc}
\begin{align}
\min _{x_{0:N}, u_{0:N-1}, \lambda_{0:N-1}} & \sum_{k=0}^{N-1} \ell(x_k,u_k) + \ell_f(x_N) \\
\text{s.t.} \quad & x_{k+1} = f(x_k, u_k, \lambda_k), \\
& 0 \leq \lambda_k \perp \Phi(x_k, u_k, \lambda_k) \geq 0, \\
& (x_k, u_k) \in \mathcal{C}
\end{align}
\end{subequations}
Here, $\ell(\cdot)$ and $\ell_f(\cdot)$ denote the stage and terminal costs, respectively, while constraints (\ref{eqn:ci_mpc}d) impose the initial condition and state/input bounds.

Solving (\ref{eqn:ci_mpc}) in real time is generally intractable even for relatively simple multi-contact systems, primarily due to the non-smoothness and combinatorial complexity of nonlinear complementarity constraints (\ref{eqn:ci_mpc}c).
As a result, many approaches resort to local approximations of the contact dynamics in (\ref{eqn:dynamics}) to obtain tractable formulations \cite{Aydinoglu2024, cleac2024}.
These approximations are often too slow for complex, multi-contact scenarios, and their more critical limitation is locality.
Local models restrict the controller to a narrow region of validity, preventing it from taking short-term suboptimal actions that are necessary for long-term success.
As a result, the controller risks becoming trapped in local minima and failing to reach distant goals.

\begin{figure}[t]
    \centering
    \includegraphics[width=\linewidth]{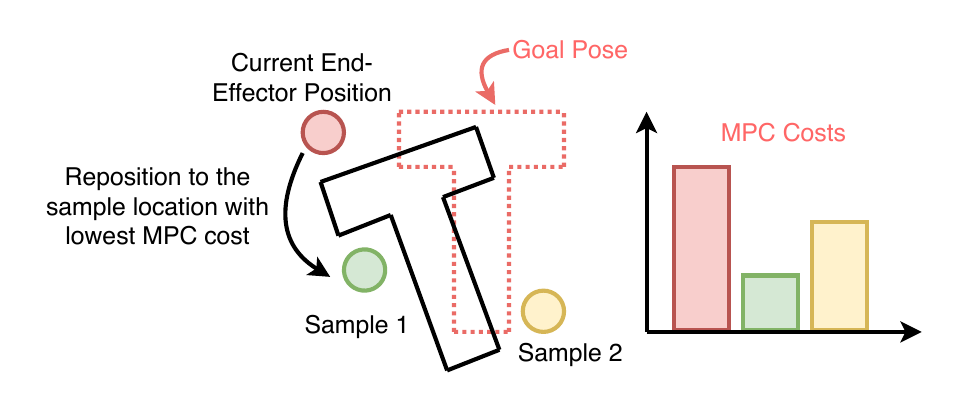}
    \caption{Illustration of sampling-based CI-MPC on the planar Push-T task \cite{chi2023diffusionpolicy}. Different end effector positions are shown with their associated MPC costs. The black solid lines denote the current pose of the T, and the red dashed lines denote the target pose. The optimal trajectory requires translating the T upward and to the right while rotating it clockwise. However, executing short-horizon MPC from the current end effector position (red circle) produces the opposite effect, pushing the T downward and counter-clockwise and resulting in higher cost. In contrast, the green candidate position yields the lowest cost, as it enables MPC to effectively align the T with its target. Thus, moving the end effector first to this green location before executing MPC facilitates more effective manipulation.}
	\label{img.sampling_based_mpc}
    \vspace{-0.5cm}
\end{figure}
\subsection{Approximate Global CI-MPC via Sampling}\label{subsec:bg_sampling_cimpc}
Overcoming locality requires global guidance, and one way of achieving this is through sampling, as in Venkatesh, Bianchini et al.~\cite{Venkatesh2025}.
In this approach, candidate end effector positions are sampled and evaluated by solving the CI-MPC problem, augmented with a travel cost from the current location. The candidate with the lowest overall cost is selected, and if it differs from the current pose, the robot first moves along a collision-free path to that position before executing CI-MPC.
As illustrated in Fig.~\ref{img.sampling_based_mpc}, by steering the system toward configurations that expand the reach of local MPC, this strategy helps the controller overcome local minima and achieve goals that require long-horizon reasoning.

\section{Methods}
We present the \textit{Push Anything} framework (Fig. \ref{img.diagram}), a pipeline integrating object perception with a novel controller.
Our framework operates in two phases.
In the offline phase, we build an object library by scanning objects to generate meshes and URDFs, assuming the same mass and inertia.
In the online phase, our controller uses robot and object state estimates to compute end effector trajectories. Following the approach in \cite{Yang2024}, these trajectories are tracked by a low-level operational space controller (OSC) \cite{Khatib1987AUA}.
We present our perception pipeline for mesh reconstruction and object tracking (\S \ref{subsec.mesh_scanning}) followed by our controller (\S \ref{subsec.controller}), highlighting two key components: the end effector sampling strategy (\S \ref{subsec.sampling_strategy}) and the improved local CI-MPC (\S \ref{subsec.c3plus}).

\subsection{Mesh Reconstruction and Object Tracking}\label{subsec.mesh_scanning}
\subsubsection{Novel Object Mesh Reconstruction} Given a new object, we record a video of it with a RealSense D455 RGBD camera.
We manually select an object mask in the first frame, and XMem \cite{cheng2022xmem} generates subsequent object masks automatically. From RGB images, depth images, masks, and camera intrinsics, BundleSDF \cite{wen2023bundlesdf} performs mesh reconstruction.
We post-process the mesh to make it watertight and orient its $z$-axis to be upwards when the object lies flat on the table.

\subsubsection{Multi-Object Tracking}
To track multiple objects, we run multiple instances of FoundationPose \cite{foundationposewen2024} in parallel, directly sharing memory access to the camera frames.
By default, FoundationPose initializes object tracking by registering the object mask in the first frame and subsequently propagating poses frame by frame. While effective in short sequences, this approach is insufficient for our setting, which involves frequent occlusions from other objects and the robot end effector, imperfect object meshes, and the need for long-horizon, high-accuracy tracking. To increase the tracking robustness, we integrate XMem \cite{cheng2022xmem} to enable periodic re-registration of the mask, thereby correcting drift accumulated over time.
We must also account for pose ambiguity, where multiple poses correspond to the same physical configuration (e.g., a symmetric flat object whose z-axis can point up or down).
To resolve this, we detect and correct sudden, implausibly large changes in orientation between consecutive timesteps by selecting the pose that maintains temporal consistency.

\subsection{Sampling-Based CI-MPC Controller}\label{subsec.controller}
The success of the sampling-based CI-MPC framework in \cite{Venkatesh2025} hinges on two key components: a robust sampling strategy for selecting candidate end effector locations, and a fast, effective local CI-MPC.
While this approach is effective, applying it to arbitrary objects, particularly in multi-object scenarios, requires significant adaptations. We build upon the original framework by introducing targeted improvements to both the sampling strategy and the local CI-MPC.

\subsubsection{Sampling Strategy of End Effector Positions}\label{subsec.sampling_strategy}
\begin{figure}[t]
    \centering
    \includegraphics[width=\linewidth]{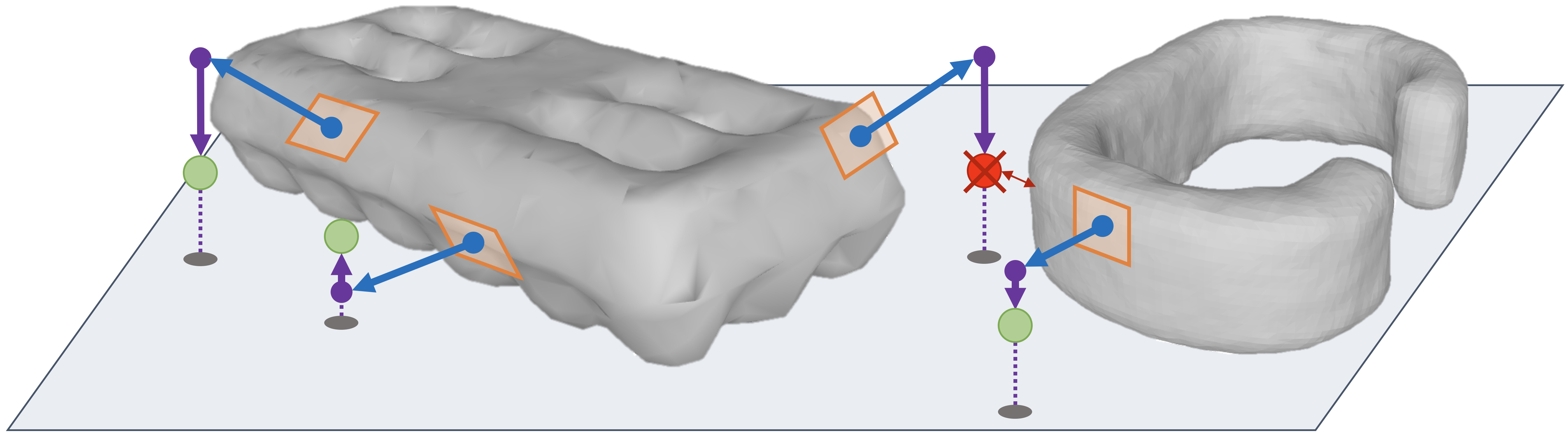}
    \caption{Visualization of the sampling strategy for end effector locations. The gray plane indicates the ground, and the orange planes represent local tangent planes to the mesh surfaces.  Blue arrows project surface samples outwards from the mesh along the face normals, then purple arrows project those to a fixed height in the world, generating candidate samples (green dots). Samples too close to object surfaces (e.g. red dot) are discarded.}
\label{fig:sampling_method}
\vspace{-0.5cm}
\end{figure}

We pre-process object meshes by storing body-frame face locations, areas, and normal vectors.  Given world-frame object pose estimates, we generate a candidate end effector location by first performing several random sampling steps in series:  1) select an object uniformly, 2) select a stored face of this object weighted by area, and 3) sample a point lying on this face.
This surface point is then projected a fixed distance along the face's outward normal vector then projected to a fixed world height (Fig. \ref{fig:sampling_method}).
We reject samples which, even after projection away from one face, are too close to any of the objects.  This can occur due to object non-convexity, the presence of multiple objects, or the selection of a face whose normal is too vertical.  We repeat this process until the desired number of end effector candidates is obtained.
\subsubsection{Consensus Complementarity Control Plus (C3+)}\label{subsec.c3plus}
For each candidate end effector location, we solve a local CI-MPC and select the lowest-cost solution. To do so efficiently, we approximate (\ref{eqn:dynamics}) by linearizing $f$ and $\Phi$ with respect to $x$, $u$, and $\lambda$, where $x$ comprises of the current object state and the sampled end effector location.
The result is a Linear Complementarity System (LCS) with dynamics:
\begin{subequations}
\label{eqn:lcs}
    \begin{align}
& x_{k+1} = A x_k + B u_k + D \lambda_k + d, \\
& 0 \leq \lambda_k \perp E x_k + F \lambda_k + H u_k + c \geq 0, \label{eqn:lcs:b}
\end{align}
\end{subequations}
where
$A \in \mathbb{R}^{n_x \times n_x}, B \in \mathbb{R}^{n_x \times n_u}, D \in \mathbb{R}^{n_x \times n_\lambda}, d \in \mathbb{R}^{n_x}, E \in \mathbb{R}^{n_\lambda \times n_x}, F \in \mathbb{R}^{n_\lambda \times n_\lambda}, H \in \mathbb{R}^{n_\lambda \times n_u}$, and $c \in \mathbb{R}^{n_\lambda}$.
While using linearized terms, this model preserves the multi-modal nature of contact dynamics through the complementarity constraint (\ref{eqn:lcs:b}).

Combining this LCS model with a standard quadratic cost function yields a Quadratic Program with Complementarity Constraints (QPCC), a well-known class of non-convex optimization problems that can be reformulated into a Mixed-Integer Quadratic Program (MIQP) \cite{Bai2012OnCQ}:
\begin{subequations}
\label{eqn:ci_mpc_with_lcs}
\begin{align}
\min_{\substack{x_{0:N}, u_{0:N-1} \\ \lambda_{0:N-1}}} \;&  
\sum_{k=0}^{N-1}\left(x_k^T Q_k x_k+u_k^T R_k u_k\right)+x_N^T Q_N x_N \\
\text {s.t. } \;&
x_{k+1}=A x_k+B u_k+D \lambda_k+d, \\
& 0 \leq \lambda_k \perp E x_k+F \lambda_k+H u_k+c \geq 0, \\
& (x_k, u_k) \in \mathcal{C}.
\end{align}
\end{subequations}
Our method, C3+, seeks a more efficient solution than solving with an MIQP.
C3+ builds upon the Consensus Complementarity Control (C3) framework \cite{Aydinoglu2024}, which employs the Alternating Direction Method of Multipliers (ADMM) \cite{boyd2011}.
The key insight of C3+ is to reformulate the problem (\ref{eqn:ci_mpc_with_lcs}) by introducing a slack variable, $\eta_k \in \mathbb{R}^{n_{\lambda}}$, to represent the linear expression within the complementarity constraint. By defining $\eta_k = E x_k + F \lambda_k + H u_k + c$, we arrive at the following equivalent problem:
\begin{subequations}
\label{eqn:ci_mpc_with_lcs_with_slack}
\begin{align}
\min_{\substack{x_{0:N}, u_{0:N-1} \\ \lambda_{0:N-1} \\ \eta_{0:N-1}}} \;&  
\sum_{k=0}^{N-1}\left(x_k^T Q_k x_k+u_k^T R_k u_k\right)+x_N^T Q_N x_N \label{eqn:ci_mpc_with_lcs_with_slack:a} \\
\text {s.t. } \;&
x_{k+1}=A x_k+B u_k+D \lambda_k+d, \label{eqn:ci_mpc_with_lcs_with_slack:b} \\
& \eta_k = E x_k+F \lambda_k+H u_k+c, \label{eqn:ci_mpc_with_lcs_with_slack:c} \\
& 0 \leq \lambda_k \perp \eta_k \geq 0, \label{eqn:ci_mpc_with_lcs_with_slack:d} \\
& (x_k, u_k) \in \mathcal{C}. \label{eqn:ci_mpc_with_lcs_with_slack:e}
\end{align}
\end{subequations}
To apply the ADMM-based strategy, C3+ first reframes (\ref{eqn:ci_mpc_with_lcs_with_slack}) into a consensus form.
This leads to an augmented decision variable $z_k^T = [x_k^T, \lambda_k^T, u_k^T, \eta_k^T]$. We then create a copy, $\delta_k^T = [\left(\delta^x_k\right)^T, \left(\delta^{\lambda}_k\right)^T, \left(\delta^u_k\right)^T, \left(\delta^\eta_k\right)^T]$, of this variable, allowing us to split the constraints into two sets and rewrite (\ref{eqn:ci_mpc_with_lcs_with_slack}) as:
\begin{equation}
\label{eqn:consensus_form}
\begin{array}{ll}
     \min\limits_{z} & c(z) + \mathcal{I}_{\mathcal{D}}(z) + \displaystyle \sum_{k=0}^{N-1}\mathcal{I}_{\mathcal{H}_k}(\delta_k) \\
     \text{s.t.} & z_k = \delta_k, \quad \forall k=0, \dots, N-1.
\end{array}    
\end{equation}
Here, $z^T = [z_0^T,z_1^T,...,z^T_{N-1}]$, $\delta^T=[\delta_0^T,\delta_1^T,...,\delta^T_{N-1}]$, $c(z)$ is the original quadratic cost in \eqref{eqn:ci_mpc_with_lcs_with_slack:a}, $\mathcal{I}$ is the 0-$\infty$ indicator function, and the constraints are encoded in the sets.
The set $\mathcal{D}$ comprises all feasible $z$ satisfying the coupled constraints across time: the linear dynamics \eqref{eqn:ci_mpc_with_lcs_with_slack:b}, the slack-variable equality \eqref{eqn:ci_mpc_with_lcs_with_slack:c}, and initial and state/input bounds \eqref{eqn:ci_mpc_with_lcs_with_slack:e}. The set $\mathcal{H}_k$, by contrast, contains all feasible $\delta$ satisfying only the now-simplified complementarity constraint, $0 \leq \delta_k^{\lambda} \perp \delta_k^{\eta} \geq 0$, which is local to each timestep $k$.

The ADMM algorithm solves this consensus problem by iteratively minimizing the general augmented Lagrangian $\mathcal{L}_\rho(z, \delta, w)$ of \eqref{eqn:consensus_form} (see \cite{Aydinoglu2024} \S V-B.2), which involves cyclically performing three updates:
\begin{enumerate}
    \item Quadratic Step ($z$-update): 
    \begin{equation}
        z^{i+1} = \operatorname{argmin}_z \mathcal{L}_\rho\left(z, \delta^i, w^i\right)
    \end{equation}
    \item Projection Step ($\delta$-update):
    \begin{equation}
        \delta_k^{i+1} = \operatorname{argmin}_{\delta_k} \mathcal{L}_\rho\left(z_k^{i+1}, \delta_k, w_k^i\right), \forall k
    \end{equation}
    \item Dual Update ($w$-update):
    \begin{equation}
        w_k^{i+1} = w_k^i + z_k^{i+1} - \delta_k^{i+1}, \forall k
    \end{equation}
\end{enumerate}
where $w_k$ are the scaled dual variables and $\rho$ is a penalty parameter.
Each iteration begins with the quadratic step, updating $z$ by solving a convex Quadratic Program (QP) derived from the augmented Lagrangian:
\begin{equation}
     \label{eqn:qp_formulation_in_quadratic_step}
     \begin{aligned}
         \min_z \;& c(z) + \sum_{k=0}^{N-1} \rho \left\| z_k - \delta_k^i + w_k^i\right\|^2_G \\
         \text{s.t. } & z \in \mathcal{D}.
     \end{aligned}
\end{equation}
\begin{figure}[t]
    \centering
    \includegraphics[width=\linewidth]{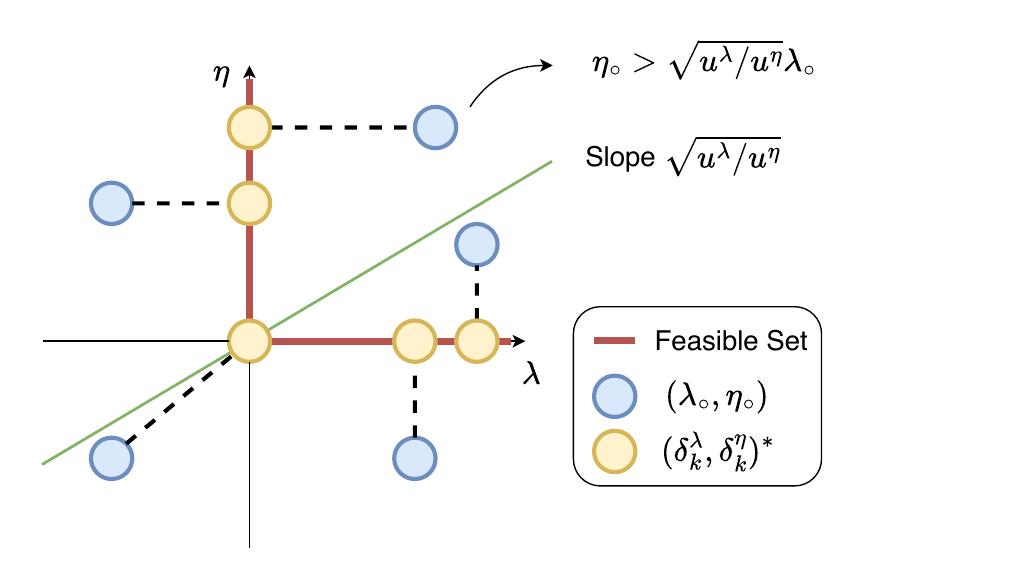}
    \caption{The C3+ projection step maps each point $(\lambda_{\circ}, \eta_{\circ})$ (blue) to its closest point $(\delta_k^\lambda, \delta_k^\eta)^\ast$ (yellow) on the feasible complementarity set (red axes). Points in the positive orthant are projected onto a nearby axis, with projection weights forming a linear switching boundary (green).}
    \label{fig:c3plus_projection_step}
    \vspace{-0.5cm}
\end{figure}
Unlike C3, C3+ augments the set $\mathcal{D}$ with an additional linear equality constraint on $\eta_k$ as given in \eqref{eqn:ci_mpc_with_lcs_with_slack:c}.
While this slightly increases the size of the QP, it remains convex and can be solved efficiently by standard solvers like OSQP \cite{osqp}.

The solution of this QP, $z^{i+1}$, then feeds into the second step, the projection step. This operation decouples across all timesteps $k$, allowing for parallel computation. For each timestep, the task is to project the output from the first step onto the simple complementarity set $\mathcal{H}_k$:
\begin{equation}
\label{eqn:projection_step}
\begin{array}{cl}
\min\limits_{\delta_k} & \|\delta_k - (z_k^{i+1} + w_k^i)\|^2_U \\
\text{s.t. } & 0 \leq \delta_k^{\lambda} \perp \delta_k^{\eta} \geq 0.
\end{array}
\end{equation}

\begin{figure}[t]
    \centering
    \includegraphics[width=0.7\linewidth]{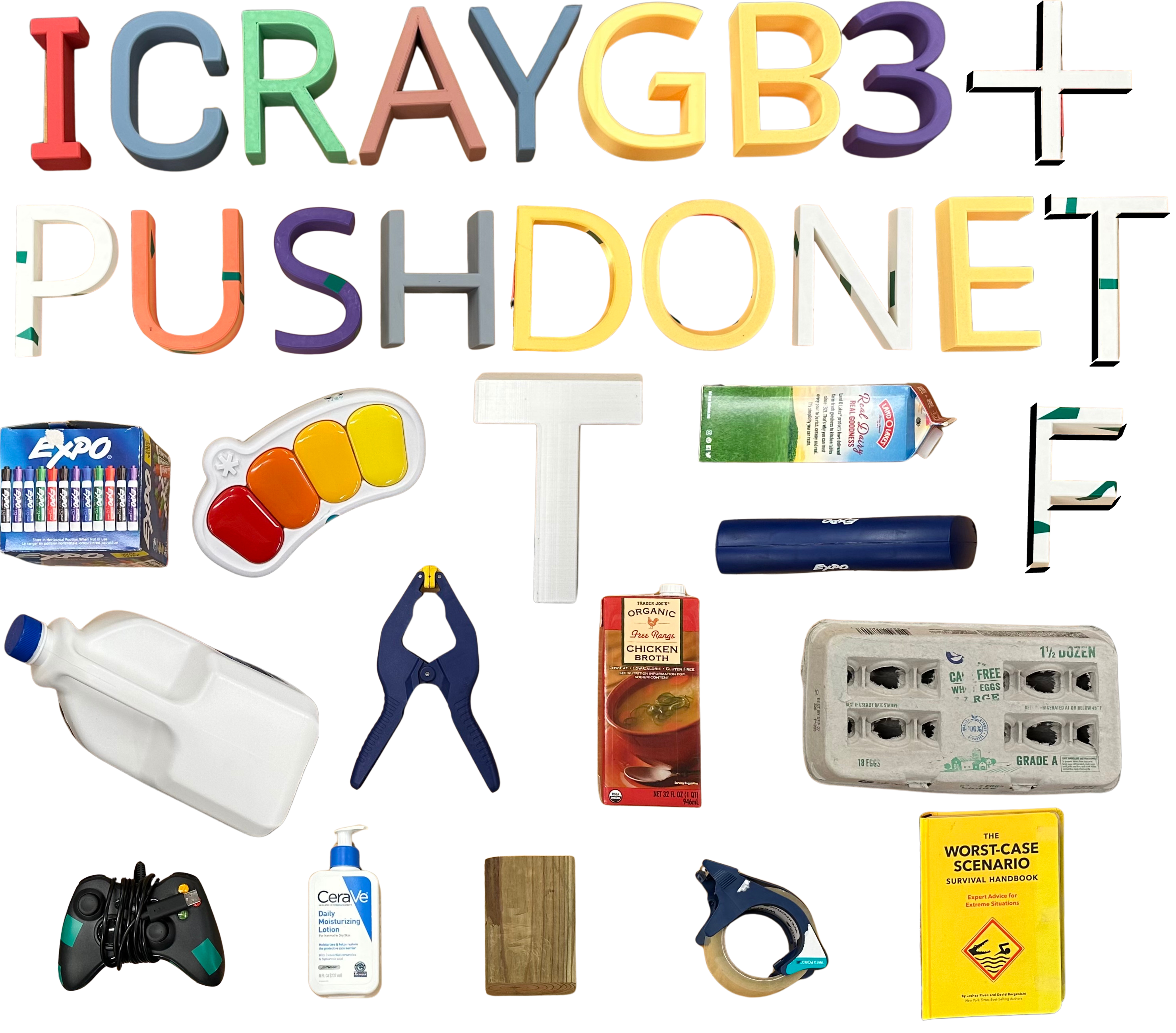}
    \caption{Diverse objects in Push Anything hardware experiments.}
	\label{fig:objects}
\end{figure}
\begin{figure}[t]
    \centering
    \includegraphics[width=0.7\linewidth]{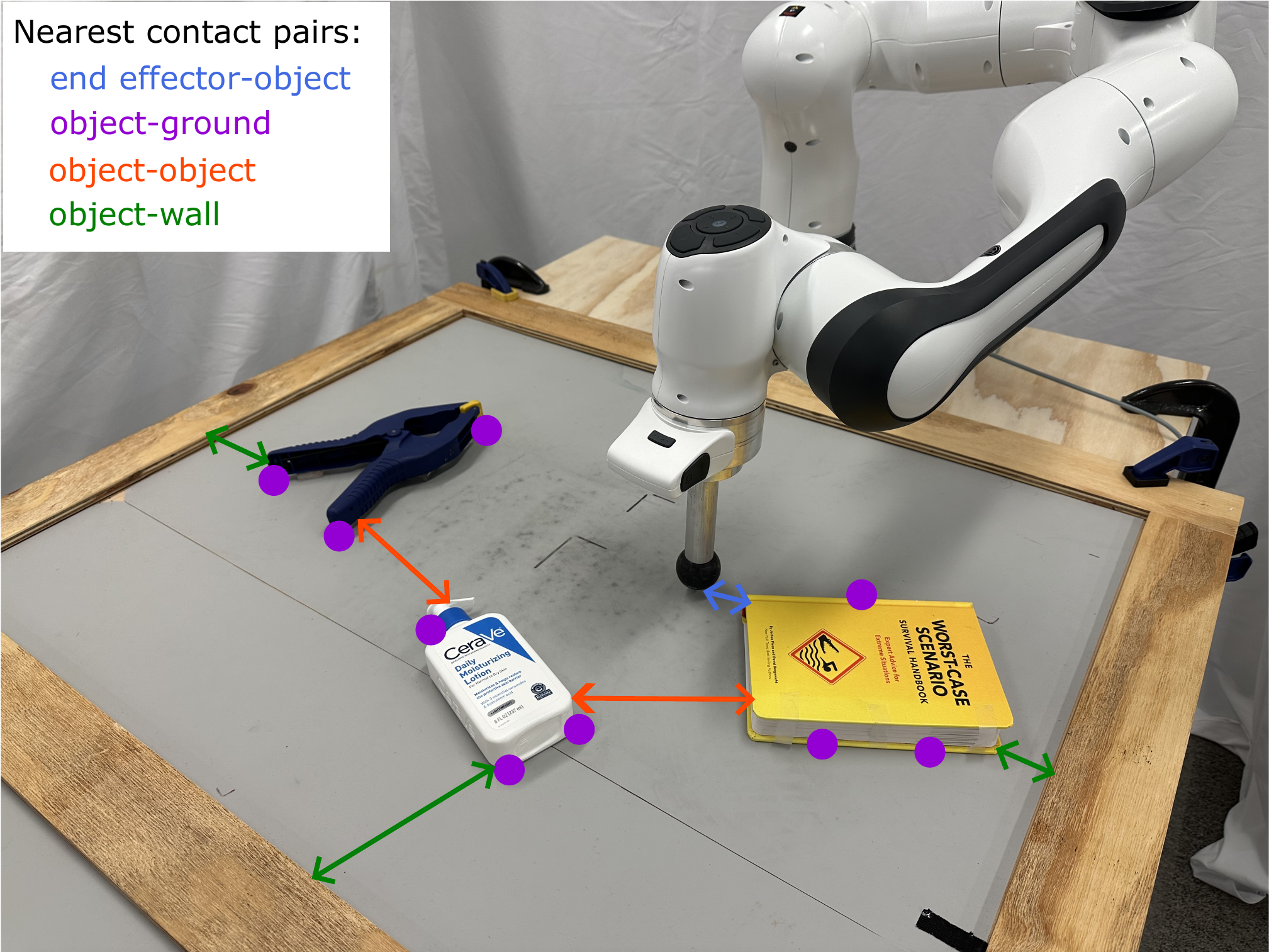}
    \caption{Visualization of the selected contact pairs in planar pushing task.}
	\label{fig:contact_modelling}
    \vspace{-0.5cm}
\end{figure}
This projection is the source of C3+'s significant computational advantage.
The introduction of a slack variable means that the non-convex component, the complementarity constraint, becomes decoupled across contacts. Therefore, solving \eqref{eqn:projection_step} is equivalent to solving many independent 1D MIQPs, each of which has a simple closed-form solution, effectively transforming the costly, coupled exponential-time MIQP into a constant-time analytical computation and yielding a significant overall speedup.
As defined below and illustrated in Fig. \ref{fig:c3plus_projection_step}, the optimal value for each component of $(\delta_k^\lambda, \delta_k^\eta)$ is computed as
\begin{equation}
\label{eqn:closed_form_sol_projection}
(\delta_k^\lambda, \delta_k^\eta)^\ast =
\begin{cases}
(0, \eta_\circ) & \text{if } \eta_\circ \ge 0 \text{ and } \eta_\circ \ge \sqrt{u^\lambda / u^\eta}\, \lambda_\circ, \\
(\lambda_\circ, 0) & \text{if } \lambda_\circ \ge 0 \text{ and } \eta_\circ < \sqrt{u^\lambda / u^\eta}\, \lambda_\circ, \\
(0,0) & \text{otherwise,}
\end{cases}
\end{equation}
where $\lambda_\circ = (z_k^{i+1} + w_k^i)^\lambda, \eta_\circ = (z_k^{i+1} + w_k^i)^\eta$ and $u^\lambda, u^\eta > 0$ are the weights. Note that the above formula is applied \emph{element-wise} for each component of $(\delta_k^\lambda, \delta_k^\eta)$.
After the projection is complete, the third and final step of the iteration is the dual update, where the scaled dual variables $w_k$ are adjusted to drive the consensus variables $z_k$ and $\delta_k$ toward agreement in the next iteration.
\begin{figure*}[h]
    \centering
    \includegraphics[width=0.97\textwidth]{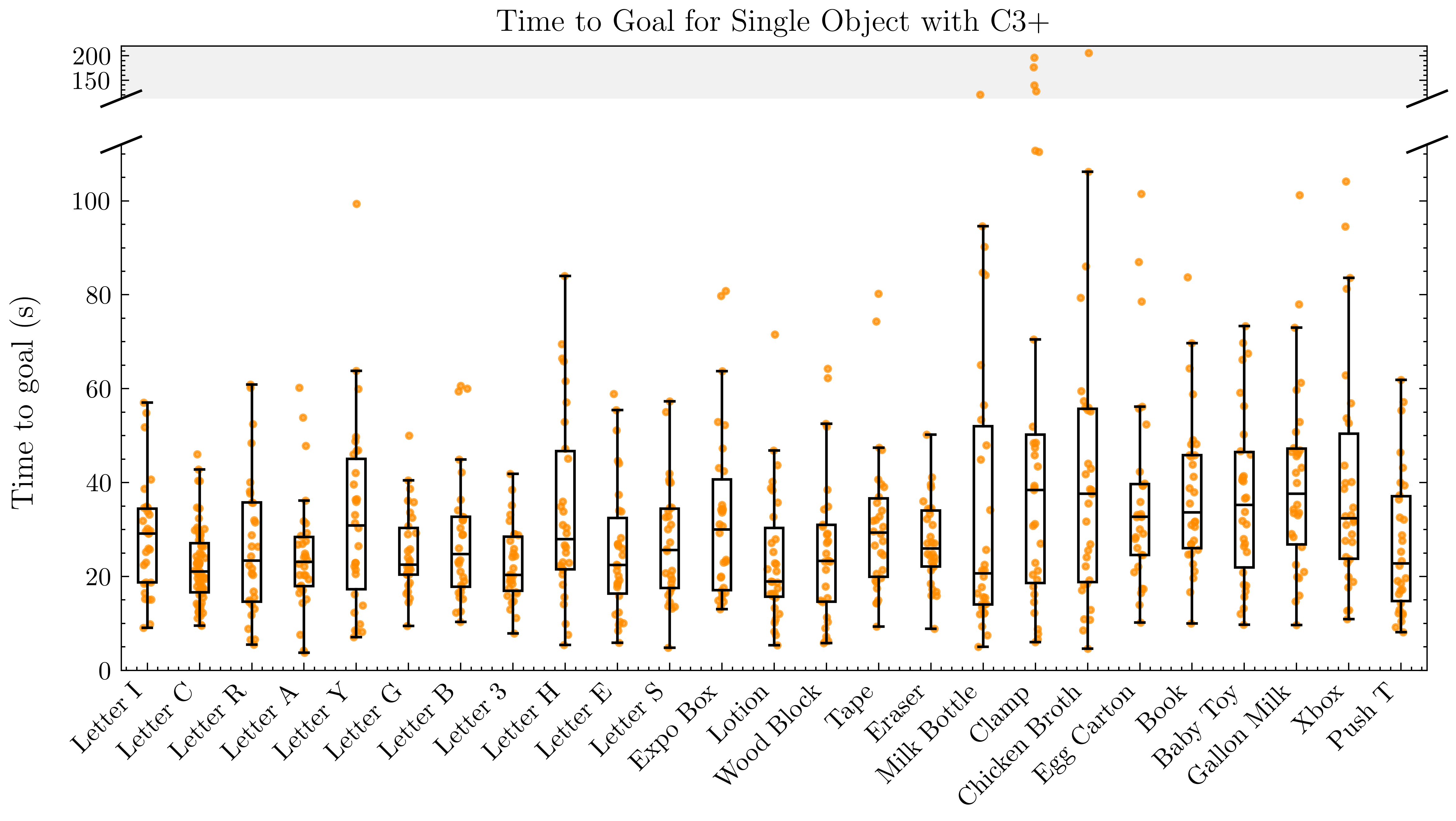}
    \vspace{-0.3cm}
    \caption{Time-to-goal distributions for various objects. Boxplots show the median and interquartile range, while orange dots represent individual data points from each trial. For visualization clarity, the y-axis is truncated. The shaded region notes the presence of outliers that fall beyond this truncated range.}
	\label{fig:time_to_goal}
    \vspace{-0.45cm}
\end{figure*}

Although many ADMM iterations could be performed to achieve full convergence, in practice we terminate early after a small, fixed number of iterations, accepting a potentially suboptimal but sufficiently good solution, to maintain real-time control rates.
Additionally, we terminate after the quadratic step, as empirical observations indicate this yields better performance.
In the final quadratic step, we set large weights, 1000 in our implementation, in the weight matrix $G$ of the augmented cost \eqref{eqn:qp_formulation_in_quadratic_step} for the end effector–object contact components of $z^{\lambda}_k$ and $z^{\eta}_k$.
This encourages the end effector–object forces to closely match their values from the previous projection step, which satisfies the complementarity constraints and results in more dynamically feasible forces.
While this could theoretically be applied to all contacts, in practice it is too restrictive and can cause QP solver failures.

We note that the concurrent work \cite{menager:hal-05201780} independently developed an approach similar to C3+, though in the context of an inverse dynamics controller.

\section{Hardware Experiments}
\subsection{Experimental Setup}
\subsubsection{Task Description}
To evaluate our pipeline, we conduct experiments on a Franka Panda arm equipped with a spherical end effector, tasked with manipulating objects to random pose goals.
Once objects reach their goals within specified position and orientation tolerances, new pose goals are generated, thereby demonstrating generalization across diverse initial and goal poses.
We select a diverse set of 33 objects including convex and non-convex shapes, from 3D-printed letters to household objects (Fig. \ref{fig:objects}).

\begin{table}[t]
\centering
\renewcommand{\arraystretch}{1.2}
\begin{tabular}{|c|c|c|c|c|c|c|}
\hline
\textbf{\# Objs} & \textbf{\bm{$n_x$}} & \textbf{\bm{$n_{\lambda}$}} & \textbf{N} & \textbf{\bm{$\Delta$}t (s)} & \textbf{\# ADMM} & \textbf{\# Samples} \\ \hline
1 & 19 & 20 & 10 & 0.075 & 3& 6 \\ \hline
2 & 32 & 40 & 15 & 0.075 & 3& 5\\ \hline
3 & 45 & 64 & 7 & 0.075 & 3& 5\\ \hline
4 & 58 & 76 & 7 & 0.075 & 3& 5\\ \hline
\end{tabular}
\caption{Controller settings for 1-4 object experiments.}
\label{tab:controller_params}
\vspace{-1.0cm}
\end{table}

\begin{table*}[t]
\centering
\begin{tabular}{|c|c|c|c|c|c|c|c|}
\hline
\multirow{3}{*}{\textbf{\# Objs}} & \multirow{3}{*}{\textbf{Object Names}} & \multirow{3}{*}{\parbox{0.8cm}{\centering\textbf{Success Rate}}} & \multirow{3}{*}{\parbox{1.5cm}{\centering\textbf{Control Rate (Hz)}}} & \multicolumn{4}{c|}{\textbf{Time to Goal (s) within Pose Tolerances}} \\ \cline{5-8}
&&&& \multicolumn{2}{c|}{Tight (2cm, 0.1rad)} & \multicolumn{2}{c|}{Loose (5cm, 0.4rad)} \\ \cline{5-8}
&&&& $\text{Mean} \pm \sigma$ & Min, Max & $\text{Mean} \pm \sigma$ & Min, Max  \\
\hline
\multirow{10}{*}{2}
 & Lotion \& Letter R & \multirow{10}{*}{\parbox{0.8cm}{\centering100/102 (98.0\%)}} & 14.06 & $97.01 \pm 45.92$ & $47.49, 204.35$ & $74.26 \pm 39.33$ & $17.91, 171.49$ \\ \cline{2-2}\cline{4-8}
 & Baby toy \& Letter E && 14.34 & $106.91 \pm 47.62$ & $60.98, 232.75$ & $54.13 \pm 14.28$ & $20.95, 71.86$ \\ \cline{2-2}\cline{4-8}
 & Letter B \& Letter 3 && 14.31 & $85.31 \pm 28.30$ & $41.97, 129.53$ & $53.64 \pm 24.39$ & $17.98, 114.74$ \\ \cline{2-2}\cline{4-8}
 & Chicken Broth \& Expo Box && 13.84 & $99.35 \pm 38.26$ & $63.78, 180.85$ & $75.94 \pm 28.79$ & $53.77, 158.16$ \\ \cline{2-2}\cline{4-8}
 & Chicken Broth \& Wood Block && 14.12 & $74.43 \pm 20.54$ & $41.69, 107.64$ & $54.62 \pm 14.29$ & $32.24, 76.12$ \\ \cline{2-2}\cline{4-8}
 & Clamp \& Letter I && 14.14 & $87.19 \pm 26.44$ & $42.45, 132.87$ & $44.15 \pm 15.77$ & $28.52, 80.74$ \\ \cline{2-2}\cline{4-8}
 & Book \& Letter S && 14.43 & $92.88 \pm 27.27$ & $67.60, 168.00$ & $81.51 \pm 27.38$ & $61.09, 154.14$ \\ \cline{2-2}\cline{4-8}
 & Tape \& Letter A && 14.36 & $119.07 \pm 44.10$ & $69.88, 231.70$ & $73.62 \pm 24.24$ & $50.63, 138.86$ \\ \cline{2-2}\cline{4-8}
 & Letter T \& Letter H && 11.76 & $79.63 \pm 21.47$ & $44.14, 119.52$ & $47.17 \pm 11.19$ & $34.06, 69.34$ \\ \cline{2-2}\cline{4-8}
 & Letter G \& Xbox && 13.93 & $104.37 \pm 23.78$ & $75.97, 149.74$ & $64.50 \pm 14.89$ & $45.15, 89.71$ \\ \hline
\multirow{6}{*}{3}
 & Letter R \& Letter A \& Letter S & \multirow{6}{*}{\parbox{0.8cm}{\centering60/62 (96.8\%)}} & 14.70 & $185.94 \pm 63.01$ & $113.44, 299.09$ & $141.28 \pm 48.89$ & $94.30, 227.94$ \\ \cline{2-2}\cline{4-8}
 & Letter C \& Letter 3 \& Letter + && 14.85 & $173.43 \pm 29.56$ & $135.00, 219.33$ & $116.54 \pm 21.72$ & $80.66, 148.72$ \\ \cline{2-2}\cline{4-8}
 & Letter A \& Letter N \& Letter Y && 14.72 & $159.63 \pm 37.27$ & $111.57, 237.65$ & $121.65 \pm 26.65$ & $92.85, 175.08$ \\ \cline{2-2}\cline{4-8}
 & Letter I \& Letter N \& Letter G && 14.84 & $173.98 \pm 46.20$ & $115.48, 275.02$ & $120.18 \pm 40.28$ & $68.20, 189.52$ \\ \cline{2-2}\cline{4-8}
 & Letter D \& Letter I \& Letter Y && 14.72 & $188.96 \pm 38.93$ & $139.52, 247.20$ & $137.41 \pm 19.84$ & $106.46, 173.66$ \\ \cline{2-2}\cline{4-8}
 & Clamp \& Lotion \& Book && 15.19 & $224.03 \pm 53.01$ & $148.22, 307.58$ & $157.98 \pm 41.55$ & $97.30, 228.89$ \\ \cline{4-8}
\hline
\multirow{5}{*}{4}
 & PUSH & \multirow{5}{*}{\parbox{0.8cm}{\centering 50/63 (79.3\%)}} &9.32 & $312.34 \pm 60.07$ & $160.34, 394.28$ & $248.64 \pm 60.06$ & $159.82, 366.18$ \\ \cline{2-2}\cline{4-8}
 & ICRA && 8.63 & $267.85 \pm 59.80$ & $176.62, 396.33$ & $208.86 \pm 44.50$ & $126.65, 278.78$ \\ \cline{2-2}\cline{4-8}
 & URDF && 9.10 & $269.01 \pm 93.90$ & $149.50, 465.47$ & $204.92 \pm 84.96$ & $121.98, 401.88$ \\ \cline{2-2}\cline{4-8}
 & C3PO && 9.31 & $281.66 \pm 85.97$ & $120.23, 458.00$ & $192.54 \pm 38.75$ & $120.23, 246.98$ \\ \cline{2-2}\cline{4-8}
 & DAY+ && 9.12 & $326.67 \pm 117.10$ & $202.43, 597.04$ & $213.84 \pm 58.18$ & $149.72, 333.82$ \\ \cline{2-2}\cline{4-8}
\hline
\end{tabular}
\caption{Multi-object pushing performance across object sets.}
\label{tab:multi_object_result}
\vspace{-0.7cm}
\end{table*}

\subsubsection{System State and Contact Modeling}
The state $x$ includes the end effector position, the positions and orientations of the objects, as well as all velocities.
The control input $u \in \mathbb{R}^3$ represents the Cartesian forces applied at the end effector. Contact forces are captured by $\lambda \in \mathbb{R}^{4n_c}$, where $n_c$ is the number of contact pairs, each approximated using a 4-sided polyhedral friction cone \cite{Anitescu1997}.

We predefine contact geometries, but contact point pairs and their corresponding normals are determined dynamically via collision detection at each control loop.
In our setup, we define: one contact pair between the end effector and its nearest object (blue arrow in Fig. \ref{fig:contact_modelling}), three contact pairs for each object with the ground (purple circles), one contact pair for each object with the wall (green arrow, omitted in the 4-object setting), and one contact pair between every pair of objects.
In the 3-object setting shown in Fig. \ref{fig:contact_modelling}, this results in 16 contact pairs, yielding $\lambda \in \mathbb{R}^{64}$. The 1-, 2-, and 4-object settings involve 5, 10, and 19 contact pairs, respectively.

\subsubsection{Controller Parameters}
Table \ref{tab:controller_params} lists the key controller parameters: state and contact force dimensions, the planning horizon $N$, timestep $\Delta t$, ADMM iterations, and the number of end effector position samples used per control loop.

\subsection{Implementation}
The controller is implemented in C++ using Drake's Systems framework \cite{drake}. We use three computers in our experiments: (i) an Intel Core i9-13900KF (13th-gen, 32 threads) dedicated to the sampling-based controller, (ii) an Intel Core i7-9700K running the operational-space controller and robot drivers on a real-time kernel for Franka communication, and (iii) an Intel Core i9-14900K paired with an NVIDIA GeForce RTX 4090 for FoundationPose \cite{foundationposewen2024}. All computers communicate via LCM\cite{5649358}.

\subsection{Results For Single-Object Pushing}
As detailed in Fig. \ref{fig:time_to_goal}, we evaluated our method in 701 hardware trials, testing 25 objects, with each object run until 28 successful trials were obtained.
The system achieved a 99.9\% success rate (700/701), with the only failure occurring when the large egg carton was pushed out of the robot's reach.
The mean time-to-goal across all trials is approximately \(31\,\mathrm{s}\), evaluated under tight success criterion requiring translational error \(\leq 2\,\mathrm{cm}\) and rotational error \(\leq 0.1\,\mathrm{rad}\) (\(5.7^\circ\)).
For the Push T task, our framework achieves a mean time-to-goal of \(26.9\,\mathrm{s}\), improving upon prior work~\cite{Venkatesh2025} at \(30.5\,\mathrm{s}\) by \(3.5\,\mathrm{s}\) (about \(11.5\%\)) while being more broadly applicable.
The few outliers for the chicken broth and milk bottle occurred when the robot took longer to bring the objects back into reach, while the clamp's numerous outliers are better explained by its difficult physical properties, such as its complex shape, greater mass, and high friction.

\subsection{Results For Multi-Object Pushing}
In the 3- and 4-object tasks, we shorten the planning horizon to maintain a real-time control rate, as the growing number of contacts would otherwise slow down computation (Table~\ref{tab:controller_params}).
While a shorter horizon can limit long-term look-ahead, it enables faster reasoning per step and higher sampling rate, yielding a practical trade-off.
With these settings, we conducted a total of 227 trials, comprising 10 experiments for the 2-object case, 6 for the 3-object, and 5 for the 4-object case.
Each experiment was run until 10 successful trials were achieved.
Under the tight tolerance, our method achieved a 92.5\% success rate (210/227), with time-to-goal statistics for both tight and loose tolerances reported in Table~\ref{tab:multi_object_result}.
All failures occurred when an object moved beyond the robot's reach.
The mean time-to-goal for these tasks was approximately 96.4s, 191.1s, and 315.7s for the 2-, 3-, and 4-object settings, respectively.
These numbers do not scale linearly with the number of objects because goal assignments are permuted across objects in each trial, requiring object rearrangements that introduce additional execution time.
\begin{table}[t]
\centering
\renewcommand{\arraystretch}{1.2}
\resizebox{\columnwidth}{!}{
\begin{tabular}{|c|c|c|c|c|c|}
\hline
\multirow{2}{*}{\textbf{\# Objs}} & \multirow{2}{*}{\textbf{Step}} 
& \multicolumn{2}{c|}{\textbf{C3} \cite{Aydinoglu2024}} & \multicolumn{2}{c|}{\textbf{C3+ (ours)}} \\ \cline{3-6}
 & & Mean $\pm$ $\sigma$ & Max & Mean $\pm$ $\sigma$ & Max \\ \hline
\multirow{2}{*}{1} & Quadratic  & 1.67 $\pm$ 0.39 & 5.45      & 3.09 $\pm$ 0.12     & 5.67    \\ \cline{2-6}
& Projection & 10.38 $\pm$ 3.84     &  41.27    & 0.007 $\pm$ 0.001 & 0.085      \\ \hline
\multirow{2}{*}{2} & Quadratic  &  3.87 $\pm $0.94 & 7.69 &   9.13 $\pm$ 0.44   &  13.50    \\ \cline{2-6}
                   & Projection &  37.2 $\pm$ 9.12  &  131.98 &   0.011 $\pm$ 0.003  &  0.043     \\ \hline
\multirow{2}{*}{3} & Quadratic  & 2.74 $\pm$ 0.47 & 5.82 &   7.97 $\pm$ 0.02  & 13.36 \\ \cline{2-6}
                   & Projection & 40.39 $\pm$ 11.17 & 1241.85 &  0.006 $\pm$ 0.001 & 0.038 \\ \hline
\multirow{2}{*}{4} & Quadratic  &  4.59 $\pm$ 0.67 & 8.56    &  10.10 $\pm$ 0.69     & 16.02    \\ \cline{2-6}
                   & Projection &  44.07 $\pm$ 11.92 &  704.23     & 0.007 $\pm$ 0.002 & 0.041 \\ \hline
\end{tabular}
}
\caption{Comparison of solve times (ms) for C3 and C3+}
\label{tab:solve_times}
\vspace{-0.9cm}
\end{table}
\subsection{Comparison of Solve Times for C3 and C3+}
We benchmark our CI-MPC algorithm C3+ against its predecessor, C3 \cite{Aydinoglu2024}, to highlight its substantial speedup. Solve times are reported for 1-, 2-, 3-, and 4-object scenarios, totaling 103,959, 42,306, 78,129, and 40,161 solves, respectively. As shown in Table \ref{tab:solve_times}, C3+ achieves faster overall performance: while the quadratic step is slightly slower, the projection step is four to five orders of magnitude faster.

\section{Limitations and Future Work}
While our framework demonstrates strong performance in both single- and multi-object settings, overall performance is constrained by the accuracy of pose tracking provided by FoundationPose, particularly in multi-object scenes.
When objects fully or partially occlude one another, per-object tracking degrades, propagating to the controller and affecting performance.
Next steps could include improving perception robustness with multi-view tracking.
Another limitation is we model all objects with identical mass and inertia. While effective for the similar objects in our experiments, scaling to greater diversity will require online model learning or adaptation \cite{Bianchini2025, Huang2024}.
Furthermore, our control approach lacks high-level planning (e.g., first move object A, then B, etc.), and thus becomes more inefficient as the task complexity grows. A clear direction for future work would be to combine our method with higher-level reasoning.
Lastly, we aim to extend the pipeline to 3D non-prehensile manipulation.

\section{Conclusion}
In this work, we introduce Push Anything, a pipeline for real-time planar pushing that integrates (i) real-world object scanning and mesh reconstruction, (ii) robust object tracking, and (iii) a controller built on the framework of \cite{Venkatesh2025} with an improved local CI-MPC, called C3+. C3+ accelerates solve times by turning the costly projection in C3 \cite{Aydinoglu2024} into a lightweight, closed-form operation. This enables long-horizon reasoning over numerous object–object and object–environment contacts and delivers reliable performance across a broad set of geometries.
Hardware experiments demonstrate high-precision, real-time control over 33 objects, achieving high success rates and low time-to-goal on complex multi-object systems previously considered intractable.

\ifanonymized
\else
\section*{Acknowledgments}
This work was supported by an NSF CAREER Award under Grant No. FRR-2238480 and the RAI Institute.
\fi




\bibliographystyle{IEEEtran}
\bibliography{references}

\begin{thebibliography}{10}
\providecommand{\url}[1]{#1}
\csname url@samestyle\endcsname
\providecommand{\newblock}{\relax}
\providecommand{\bibinfo}[2]{#2}
\providecommand{\BIBentrySTDinterwordspacing}{\spaceskip=0pt\relax}
\providecommand{\BIBentryALTinterwordstretchfactor}{4}
\providecommand{\BIBentryALTinterwordspacing}{\spaceskip=\fontdimen2\font plus
\BIBentryALTinterwordstretchfactor\fontdimen3\font minus \fontdimen4\font\relax}
\providecommand{\BIBforeignlanguage}[2]{{%
\expandafter\ifx\csname l@#1\endcsname\relax
\typeout{** WARNING: IEEEtran.bst: No hyphenation pattern has been}%
\typeout{** loaded for the language `#1'. Using the pattern for}%
\typeout{** the default language instead.}%
\else
\language=\csname l@#1\endcsname
\fi
#2}}
\providecommand{\BIBdecl}{\relax}
\BIBdecl

\bibitem{Aydinoglu2024}
A.~Aydinoglu, A.~Wei, W.-C. Huang, and M.~Posa, ``Consensus complementarity control for multicontact mpc,'' \emph{IEEE Transactions on Robotics}, vol.~40, pp. 3879--3896, 2024.

\bibitem{cleac2024}
S.~Le~Cleac'h, T.~A. Howell, S.~Yang, C.-Y. Lee, J.~Zhang, A.~Bishop, M.~Schwager, and Z.~Manchester, ``Fast contact-implicit model predictive control,'' \emph{IEEE Transactions on Robotics}, vol.~40, pp. 1617--1629, 2024.

\bibitem{kurtz2025}
V.~Kurtz, A.~Castro, A.~Özgün Önol, and H.~Lin, ``Inverse dynamics trajectory optimization for contact-implicit model predictive control,'' \emph{The International Journal of Robotics Research}, vol.~45, no.~1, p. 23–40, 2025.

\bibitem{Venkatesh2025}
S.~Venkatesh, B.~Bianchini, A.~Aydinoglu, W.~Yang, and M.~Posa, ``Approximating global contact-implicit mpc via sampling and local complementarity,'' \emph{IEEE Robotics and Automation Letters (RA-L)}, vol.~10, no.~11, pp. 12\,117--12\,124, 2025.

\bibitem{mason1986}
M.~T. Mason, ``Mechanics and planning of manipulator pushing operations,'' \emph{The International Journal of Robotics Research}, vol.~5, no.~3, pp. 53--71, 1986.

\bibitem{mason1996}
K.~M. Lynch and M.~T. Mason, ``Stable pushing: Mechanics, controllability, and planning,'' \emph{The International Journal of Robotics Research}, vol.~15, no.~6, pp. 533--556, 1996.

\bibitem{mason1998}
S.~Akella and M.~T. Mason, ``Posing polygonal objects in the plane by pushing,'' \emph{The International Journal of Robotics Research}, vol.~17, no.~1, pp. 70--88, 1998.

\bibitem{ogarFrameworkPushGraspingClutter}
M.~R. Dogar and S.~S. Srinivasa, ``A {{Framework}} for {{Push-Grasping}} in {{Clutter}},'' in \emph{Robotics: Science and Systems (RSS)}, 2011.

\bibitem{kuan-ting2016}
K.-T. Yu, M.~Bauza, N.~Fazeli, and A.~Rodriguez, ``More than a million ways to be pushed. a high-fidelity experimental dataset of planar pushing,'' in \emph{2016 IEEE/RSJ International Conference on Intelligent Robots and Systems (IROS)}, 2016, pp. 30--37.

\bibitem{yiPreciseObjectSliding2023}
X.~Yi and N.~Fazeli, ``Precise {{Object Sliding}} with {{Top Contact}} via {{Asymmetric Dual Limit Surfaces}},'' in \emph{Robotics: Science and Systems (RSS)}, 2023.

\bibitem{li2018}
J.~K. Li, W.~S. Lee, and D.~Hsu, ``Push-net: Deep planar pushing for objects with unknown physical properties,'' in \emph{Robotics: Science and Systems (RSS)}, 2018.

\bibitem{10341629}
J.~Del Aguila~Ferrandis, J.~Moura, and S.~Vijayakumar, ``Nonprehensile planar manipulation through reinforcement learning with multimodal categorical exploration,'' in \emph{2023 IEEE/RSJ International Conference on Intelligent Robots and Systems (IROS)}, 2023, pp. 5606--5613.

\bibitem{dengler2025learninggoaldirectedobjectpushing}
N.~Dengler, J.~D.~A. Ferrandis, J.~Moura, S.~Vijayakumar, and M.~Bennewitz, ``Learning goal-directed object pushing in cluttered scenes with location-based attention,'' \emph{arXiv preprint arXiv:2403.17667}, 2025.

\bibitem{Yang2024}
W.~Yang and M.~Posa, ``Dynamic on-palm manipulation via controlled sliding,'' in \emph{Robotics: Science and Systems (RSS)}, Jul. 2024.

\bibitem{menager:hal-05201780}
E.~M{\'e}nager, A.~Bambade, W.~Jallet, A.~de~Marchi, and J.~Carpentier, ``Contact-implicit inverse dynamics,'' \emph{hal-05201780v1}, 2025.

\bibitem{jiang2024contactimplicitmodelpredictivecontrol}
Y.~Jiang, M.~Yu, X.~Zhu, M.~Tomizuka, and X.~Li, ``Contact-implicit model predictive control for dexterous in-hand manipulation: A long-horizon and robust approach,'' in \emph{2024 IEEE/RSJ International Conference on Intelligent Robots and Systems (IROS)}, 2024, pp. 5260--5266.

\bibitem{gradyMPPI}
G.~Williams, P.~Drews, B.~Goldfain, J.~M. Rehg, and E.~A. Theodorou, ``Aggressive driving with model predictive path integral control,'' in \emph{2016 IEEE International Conference on Robotics and Automation (ICRA)}, 2016, pp. 1433--1440.

\bibitem{reuvenCEM}
R.~Y. Rubinstein and D.~P. Kroese, \emph{The Cross-Entropy Method}.\hskip 1em plus 0.5em minus 0.4em\relax Springer New York, NY, 2004.

\bibitem{zhang2025}
J.~Z. Zhang, T.~A. Howell, Z.~Yi, C.~Pan, G.~Shi, G.~Qu, T.~Erez, Y.~Tassa, and Z.~Manchester, ``Whole-body model-predictive control of legged robots with mujoco,'' \emph{arXiv preprint arXiv:2503.04613}, 2025.

\bibitem{li2024drop}
A.~H. Li, P.~Culbertson, V.~Kurtz, and A.~D. Ames, ``Drop: Dexterous reorientation via online planning,'' in \emph{2025 IEEE International Conference on Robotics and Automation (ICRA)}, 2025, pp. 14\,299--14\,306.

\bibitem{hess2024sampling}
A.~Hess, A.~M. K{\"u}bler, B.~Forrai, M.~Dogar, and R.~K. Katzschmann, ``Sampling-based model predictive control for dexterous manipulation on a biomimetic tendon-driven hand,'' \emph{arXiv preprint arXiv:2411.06183}, 2024.

\bibitem{howell2022predictivesamplingrealtimebehaviour}
T.~Howell, N.~Gileadi, S.~Tunyasuvunakool, K.~Zakka, T.~Erez, and Y.~Tassa, ``Predictive sampling: Real-time behaviour synthesis with mujoco,'' \emph{arXiv preprint arXiv:2212.00541}, 2022.

\bibitem{suh2025}
H.~J.~T. Suh, T.~Pang, T.~Zhao, and R.~Tedrake, ``Dexterous contact-rich manipulation via the contact trust region,'' \emph{The International Journal of Robotics Research}, vol.~0, no.~0, p. 02783649251398875, 0.

\bibitem{doi:10.1137/1.9781611970715}
D.~E. Stewart, \emph{Dynamics with Inequalities}.\hskip 1em plus 0.5em minus 0.4em\relax Society for Industrial and Applied Mathematics, 2011.

\bibitem{doi:10.1177/0278364913506757}
M.~Posa, C.~Cantu, and R.~Tedrake, ``A direct method for trajectory optimization of rigid bodies through contact,'' \emph{The International Journal of Robotics Research}, vol.~33, no.~1, pp. 69--81, 2014.

\bibitem{doi:10.1177/0278364919849235}
Z.~Manchester, N.~Doshi, R.~J. Wood, and S.~Kuindersma, ``Contact-implicit trajectory optimization using variational integrators,'' \emph{The International Journal of Robotics Research}, vol.~38, no. 12-13, pp. 1463--1476, 2019.

\bibitem{chi2023diffusionpolicy}
C.~Chi, S.~Feng, Y.~Du, Z.~Xu, E.~Cousineau, B.~Burchfiel, and S.~Song, ``Diffusion policy: Visuomotor policy learning via action diffusion,'' in \emph{Proceedings of Robotics: Science and Systems (RSS)}, 2023.

\bibitem{Khatib1987AUA}
O.~Khatib, ``A unified approach for motion and force control of robot manipulators: The operational space formulation,'' \emph{IEEE Journal on Robotics and Automation}, vol.~3, no.~1, pp. 43--53, 1987.

\bibitem{cheng2022xmem}
H.~K. Cheng and A.~G. Schwing, ``{XMem}: Long-term video object segmentation with an atkinson-shiffrin memory model,'' in \emph{ECCV}, 2022.

\bibitem{wen2023bundlesdf}
B.~Wen, J.~Tremblay, V.~Blukis, S.~Tyree, T.~Muller, A.~Evans, D.~Fox, J.~Kautz, and S.~Birchfield, ``Bundlesdf: Neural 6-dof tracking and 3d reconstruction of unknown objects,'' \emph{CVPR}, 2023.

\bibitem{foundationposewen2024}
B.~Wen, W.~Yang, J.~Kautz, and S.~Birchfield, ``{FoundationPose}: Unified 6d pose estimation and tracking of novel objects,'' in \emph{CVPR}, 2024.

\bibitem{Bai2012OnCQ}
L.~Bai, J.~E. Mitchell, and J.~S. Pang, ``On convex quadratic programs with linear complementarity constraints,'' \emph{Computational Optimization and Applications}, vol.~54, pp. 517 -- 554, 2012.

\bibitem{boyd2011}
S.~Boyd, N.~Parikh, E.~Chu, B.~Peleato, and J.~Eckstein, \emph{Distributed Optimization and Statistical Learning via the Alternating Direction Method of Multipliers}.\hskip 1em plus 0.5em minus 0.4em\relax Foundations and Trends in Machine Learning, 2011, vol.~3, no. 1 (2010) 1-122.

\bibitem{osqp}
B.~Stellato, G.~Banjac, P.~Goulart, A.~Bemporad, and S.~Boyd, ``{OSQP}: an operator splitting solver for quadratic programs,'' \emph{Mathematical Programming Computation}, vol.~12, no.~4, pp. 637--672, 2020.

\bibitem{Anitescu1997}
M.~Anitescu and F.~A. Potra, ``Formulating dynamic multi-rigid-body contact problems with friction as solvable linear complementarity problems,'' \emph{Nonlinear Dynamics}, vol.~14, no.~3, pp. 231--247, 1997.

\bibitem{drake}
R.~Tedrake and the Drake Development~Team, ``Drake: Model-based design and verification for robotics,'' 2019.

\bibitem{5649358}
A.~S. Huang, E.~Olson, and D.~C. Moore, ``Lcm: Lightweight communications and marshalling,'' in \emph{2010 IEEE/RSJ International Conference on Intelligent Robots and Systems}, 2010, pp. 4057--4062.

\bibitem{Bianchini2025}
B.~Bianchini, M.~Zhu, M.~Sun, B.~Jiang, T.~Camillo~J, and M.~Posa, ``Vysics: Object reconstruction under occlusion by fusing vision and contact-rich physics,'' in \emph{Robotics: Science and Systems (RSS)}, 2025.

\bibitem{Huang2024}
W.-C. Huang, A.~Aydinoglu, W.~Jin, and M.~Posa, ``Adaptive contact-implicit model predictive control with online residual learning,'' in \emph{IEEE International Conference on Robotics and Automation (ICRA)}, May 2024.

\end{thebibliography}

\end{document}